%% file: main.tex
\documentclass[letterpaper, 10 pt, conference]{ieeeconf}
\IEEEoverridecommandlockouts

\let\labelindent\relax
\usepackage{amsmath,amssymb,amsfonts,amsthm}
\usepackage{algorithmic}

\newtheorem{remark}{\bf Remark}
\makeatletter

\newcommand{\Rmnum}[1]{\expandafter\@slowromancap\romannumeral #1@}
\makeatother
\renewcommand{\algorithmicrequire}{\textbf{Input:}}  
 
\input{configs}

\begin{document}

\title{\LARGE \bf
SwarmPRM: Probabilistic Roadmap Motion Planning for Large-Scale Swarm Robotic Systems}

\author{Yunze Hu$^{1}$, Xuru Yang$^{1}$, Kangjie Zhou$^1$, Qinghang Liu$^1$, Kang Ding$^1$, Han Gao$^1$,\\ Pingping Zhu$^2$, and Chang Liu$^{1}$
\thanks{*This work was sponsored by Beijing Nova Program (20220484056) and the National Natural Science Foundation of China (62203018).}
\thanks{All correspondences should be sent to Chang Liu.}
\thanks{$^1$Yunze Hu, Xuru Yang, Kangjie Zhou, Qinghang Liu, Kang Ding, Han Gao, and Chang Liu are with the Department of Advanced Manufacturing and Robotics, College of Engineering, Peking University, Beijing 100871, China (hu\_yun\_ze@stu.pku.edu.cn; xuru.yang@stu.pku.edu.cn; kangjiezhou@pku.edu.cn; californium@stu.pku.edu.cn; kangding@stu.pku.edu.cn; hangaocoe@pku.edu.cn; changliucoe@pku.edu.cn).}
\thanks{$^2$Pingping Zhu is with the Department of Computer Sciences and Electrical Engineering (CSEE), 
Marshall University, Huntington, WV 25755, USA (zhup@marshall.edu).}}

\maketitle
\thispagestyle{empty}
\pagestyle{empty}

\begin{abstract}
Large-scale swarm robotic systems consisting of numerous cooperative agents show considerable promise for performing autonomous tasks across various sectors. 
Nonetheless, traditional motion planning approaches often face a trade-off between scalability and solution quality due to the exponential growth of the joint state space of robots. 
In response, this work proposes SwarmPRM, a hierarchical, scalable, computationally efficient, and risk-aware sampling-based motion planning approach for large-scale swarm robots.
SwarmPRM utilizes a Gaussian Mixture Model (GMM) to represent the swarm's macroscopic state and constructs a Probabilistic Roadmap in Gaussian space, referred to as the Gaussian roadmap, to generate a transport trajectory of GMM.
This trajectory is then followed by each robot at the microscopic stage. 
To enhance trajectory safety, SwarmPRM 
incorporates the conditional value-at-risk (CVaR) in the collision checking process to impart the property of risk awareness to the constructed Gaussian roadmap.
SwarmPRM then crafts a linear programming formulation to compute the optimal GMM transport trajectory within this roadmap. 
Extensive simulations demonstrate that SwarmPRM outperforms state-of-the-art methods in computational efficiency, scalability, and trajectory quality while offering the capability to adjust the risk tolerance of generated trajectories.
\end{abstract}

\section{Introduction}
Large-scale swarm robotic systems comprised of numerous autonomous robots hold great promise for executing diverse tasks such as surveillance \cite{keller2016coordinated},
environmental exploration \cite{mcguire2019minimal}, and search and rescue \cite{drew2021multi}. 
In recent years, there has been a surge in interest towards developing motion planning techniques for large-scale swarm robots \cite{honig2018trajectory, zhu2021adaptive, soria2021predictive}.

\begin{figure}[!t]
\centering
\includegraphics[width=.83\linewidth]{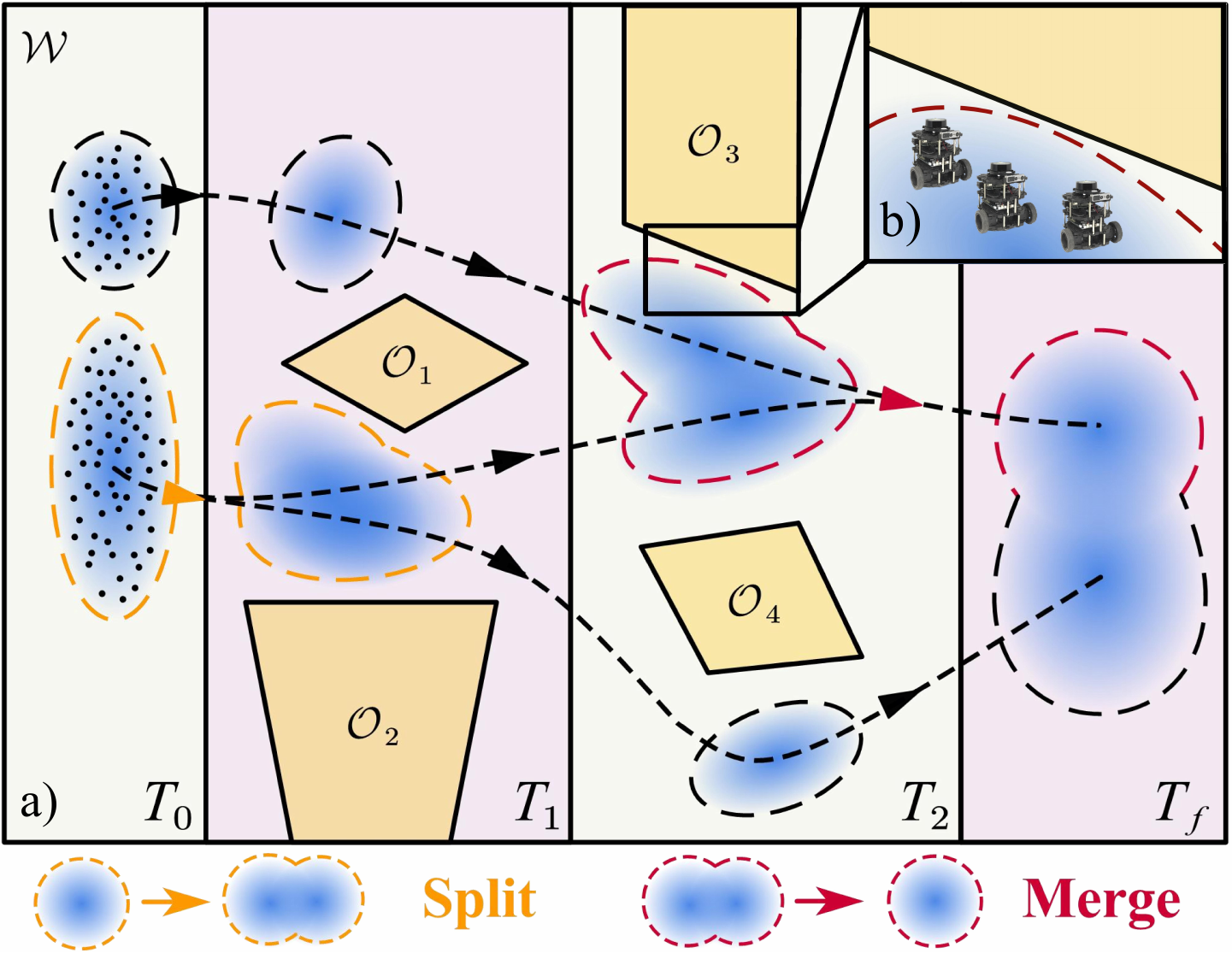}
\caption{\textbf{Illustration of the hierarchical motion planning for a large-scale robot swarm.} (a) The macroscopic planning of the swarm robots. In the two-dimensional cluttered environment $\mathcal{W}$ with four obstacles $\mathcal{O}_1,\mathcal{O}_2,\mathcal{O}_3,\mathcal{O}_4$, 
the swarm's macroscopic state, represented as a PDF, is depicted as a blue cloud, and individual robots at time $T_0$ are represented by black dots.
The macroscopic trajectory, as represented by the black dotted lines and colored arrows, guides the swarm from initial area at time $T_0$ to the target area at time $T_f$, passing through two intermediate macroscopic states at time $T_1$ and $T_2$, respectively.
The trajectory may involve "splitting" or "merging" maneuvers, depicted by the orange and red dotted lines and arrows, respectively. (b) The microscopic state of robots in the swarm. The robots track the macroscopic state while avoiding collision with obstacles.}
\label{fig:VLSR}
\end{figure}
Sampling-based algorithms \cite{karaman2011sampling} as a prominent motion planning technique have demonstrated significant potential for application in swarm robots. 
For example, Čáp et al. \cite{vcap2013multi} adapted RRT$^\ast$ for centralized multi-agent motion planning problem by treating all agents as a single system.
However, this approach is computationally intractable due to the exponential growth of the state space as the swarm size increases.  
Hönig et al. \cite{honig2018trajectory} enhanced scalability by first generating a sparse probabilistic roadmap (PRM) and then utilizing multi-agent path finding algorithms for path planning subject to inter-robot constraints. 
Yet, the sparsity of the roadmap compromises the trajectory cost in favor of computational efficiency. 
Shome et al. \cite{shome2020drrt} proposed dRRT$^\ast$, an asymptotically optimal sampling-based motion planning algorithm that constructs PRMs for individual robots and builds a search tree without requiring an explicit representation of the joint motion graph. 
Sampling-based approaches have proven effective for trajectory planning, particularly in cluttered environments. 
Nevertheless, these methods inevitably face a trade-off between scalability and the trajectory cost due to the exponential growth of the joint state space and the increasing difficulty of ensuring inter-robot constraint satisfaction.

Recently, the hierarchical strategy has emerged as a promising approach for the motion planning of large-scale swarm robots.
The hierarchical strategy consists of macroscopic and microscopic planning stages.
At the macroscopic stage, the robot swarm is treated as an entirety, and the macroscopic swarm state is usually represented as a probability density function (PDF).
At the microscopic stage, robots autonomously match the PDF by coordinating the distribution of their positions.
Adopting this strategy, Rudd et al. \cite{rudd2017generalized} employed parametrized PDFs to represent the macroscopic states of the swarm and calculated the optimal robot distribution by solving a non-convex, constrained trajectory optimization problem. 
The computational burden is very high, though, hindering its real-time implementation in practice.
Zhu et al. \cite{zhu2021adaptive} proposed an ADOC approach that modeled the macroscopic state of the swarm as a Gaussian mixture model (GMM), and utilized the optimal mass transport (OMT) theory to plan the PDF trajectory.
ADOC shows desirable real-time performance.
However, using a predetermined set of Gaussian collocation points for GMM trajectory generation limits the flexibility and adaptability to complex spatial challenges in trajectory planning, thus often yielding suboptimal and costly trajectories.

In this work, we combine the advantages of both sampling-based methods and hierarchical strategy to propose a sampling-based, hierarchical motion planning approach for robotic swarms, namely SwarmPRM.
At the macroscopic stage, the collective swarm is presented by a GMM, and a PRM is constructed to generate GMM trajectories. 
At the microscopic stage, individual robots track the PDF via a computationally efficient local tracking control law. 

A significant challenge in SwarmPRM involves collision checking during the sampling process, 
as the samples are PDFs rather than deterministic states typical of traditional sampling-based methods.
Consequently, there arises a necessity for a risk measure to assess the likelihood of collisions between a PDF sample and obstalces.
To overcome this difficulty, we propose to use the conditional value-at-risk (CVaR) as the risk measure for collision checking.
CVaR is a coherent measure \cite{artzner1999coherent} that quantifies
the potential losses beyond a certain confidence level.
Since CVaR can take into account the tail distribution through conditional expectation, which enables the discerning of rare events, it recently gained significant attention in the robotics community \cite{hakobyan2022distributionally,yang2023risk}. 

In general, SwarmPRM is a novel sampling-based, hierarchical approach for the motion planning of large-scale swarm robotic systems, 
characterized by scalability, computational efficiency, risk awareness, and trajectory flexibility.
The main contributions of this work are threefold:
\begin{enumerate}
    \item We propose to sample in Gaussian parametric space to 
    construct a Gaussian PRM where each node represents a distinct Gaussian distribution.
    The cost and geodesic path between nodes are determined under the Wasserstein metric. 
    \item We develop a systematic collision checking approach using CVaR as the risk measure when constructing the Gaussian PRM. 
    Specifically, we compute the distribution of the signed distance function (SDF) between the samples and the obstacles, and constrain the CVaR of the SDF within the safety threshold. 
    \item We design a linear programming formulation to compute the optimal macroscopic GMM trajectory on the constructed roadmap. 
    Through extensive simulations, we demonstrate the superiority of SwarmPRM over several state-of-the-art benchmark methods in aspects of computational efficiency, scalability, and trajectory quality. Furthermore, SwarmPRM exhibits risk awareness where the risk tolerance level can be easily adjusted, offering great flexibility in designing swarm behaviors,  especially in cluttered environments.
\end{enumerate}

\section{Problem Formulation and Background}
\subsection{Problem Formulation}\label{sec:ProblemFormulation}
Consider a swarm robotic system represented by the set $\mathcal{R} = \{1,2,\cdots,N\}$, where $N$ denotes the number of homogeneous robots within a large two-dimensional cluttered environment $\mathcal{W}\subset\mathbb{R}^2$ containing $N_{Obs}$ convex obstacles\footnote{Note that our motion planning method is readily adaptable to environments containing non-convex obstacles via decomposing non-convex obstacles into the union of convex ones.} $\boldsymbol{\mathcal{O}} = \{\mathcal{O}_1, \cdots, \mathcal{O}_{N_{Obs}}\}$, where $\mathcal{O}_i\subset\mathcal{W}, i=1,\cdots,N_{Obs}$. The objective is to devise trajectories for all robots, directing them from initial positions to target areas. 
The obstacles are static and known a priori, and under the assumptions of connectivity and information sharing, the states of the robots are fully observable. The set of all robots' initial positions is denoted as $\mathcal{Q}=\{\boldsymbol{q}_1,\cdots,\boldsymbol{q}_N\}$, where $\boldsymbol{q_i}\in \mathcal{W},i\in\mathcal{R}$.

We adopt a hierarchical motion planning strategy for the swarm robots (\cref{fig:VLSR}). At the macroscopic stage, we represent the entirety of the swarm as a PDF $\chi(t) \in \mathcal{P}(\mathcal{W})$ which is time-varying and constitutes a PDF trajectory with respect to time $t$. Here $\mathcal{P}(\mathcal{W})$ represents the space of PDFs with support $\mathcal{W}$. We model the initial and target distribution of swarm robots as two PDFs $\rho_{T_0}$, $\rho_{T_f}$, and aim at devising a swarm's macroscopic state trajectory transitioning from $\rho_{T_0}$ to $\rho_{T_f}$ while simultaneously avoiding obstacles. The optimal time-varying PDF $\chi(t)$ can be calculated by solving the following optimization problem
\begin{align} \label{P}
\small
\min_{\chi} \quad & J(\chi(t))\\
\textrm{s.t.}
         \quad & \rho_{T_0}=\chi(T_0), \tag{\ref{P}{a}} \label{GMM1}\\
         \quad & \rho_{T_f}=\chi(T_f), \tag{\ref{P}{b}} \label{GMM2}\\
     & R_{\alpha}(\chi(t)) \leq \delta, \forall t\in[T_0,T_f], 
     \tag{\ref{P}{c}} \label{collision}
\end{align}
where \eqref{GMM1}, \eqref{GMM2} denote the swarm's initial and target macroscopic PDF state constraints, respectively, and \eqref{collision} represents the collision avoidance constraint with $R$ and $\delta$ denoting the risk measure and safe region threshold, respectively.
The objective is to minimize the swarm's transport cost $J$, while adhering to the aforementioned constraints. At the microscopic stage, each robot tracks the PDF $\chi(t)$ while ensuring collision avoidance with obstacles and other robots.

\subsection{Sampling-Based Motion Planning Algorithms}
Sampling-based Algorithms (SBAs) have been a prevalent strategy for motion planning in robotics \cite{lavalle2006planning,karaman2011sampling}. SBAs are characterized by
a triplet $(\mathcal{X}_{free}, \boldsymbol{x}_{init}, \boldsymbol{x}_{goal})$, where $\mathcal{X}_{free}$, $\boldsymbol{x}_{init}$, and $\boldsymbol{x}_{goal}$ represent the collision-free configuration space, the initial configuration, and the goal configuration, respectively. The objective is to find a collision-free trajectory $\tau:[0,1]\rightarrow \mathcal{X}_{free}$, satisfying $\tau(0) = \boldsymbol{x}_{init}$ and $\tau(1)=\boldsymbol{x}_{goal}$. SBAs operate by sampling the configuration space and constructing a tree or graph, which expresses the connectivity and the feasible paths within the space. In this work, we specifically consider PRM \cite{karaman2011sampling}. 
Initially, new samples are drawn from $\mathcal{X}_{free}$ to create graph nodes in the roadmap. Subsequently, each node undergoes a query process to identify neighboring nodes to which the transition is collision-free with respect to obstacles, and edges connecting the respective nodes are added to the roadmap. Finally, a graph search is performed to find a shortest path connecting $\boldsymbol{x}_{init}$ and $\boldsymbol{x}_{goal}$ in the constructed roadmap.

\subsection{Optimal Transport Theory and Wasserstein Metric}
The OMT theory \cite{chen2018optimal} tackles the task of transporting masses from an initial distribution to a terminal one while maintaining mass continuity and minimizing associated costs. 
The Wasserstein distance $W_2$ is an important metric representing the minimal transport cost within the space of PDFs. 
The Wasserstein metric between two Gaussian distributions $g_1=\mathcal{N}(\boldsymbol{\mu_1},\Sigma_1), g_2=\mathcal{N}(\boldsymbol{\mu_2},\Sigma_2)$ is
\begin{equation}
\small
W_2(g_1, g_2) = \bigg \{ \Vert \boldsymbol{\mu_1}-\boldsymbol{\mu_2} \Vert^2 + tr\left[\Sigma_1+\Sigma_2-2(\Sigma_1^{\frac{1}{2}}\Sigma_2\Sigma_1^{\frac{1}{2}})^{\frac{1}{2}}\right] \bigg \}^{\frac{1}{2}},
\label{eqn:GaussianWasserstein}
\end{equation}
where $\Vert\cdot\Vert$ denotes the Euclidean distance and $tr(\cdot)$ represents the trace of a matrix. The geodesic path $\hat{g}_{1,2}(t), t\in[0,1]$ between $g_1, g_2$ is Gaussian with following mean and covariance
\begin{equation}
\small
\hat{\boldsymbol{\mu}}(t) = (1-t)\boldsymbol{\mu}_1 + t\boldsymbol{\mu}_2,
\label{eqn:interpWassersteinMean}
\end{equation}
\begin{equation}
\small
\hat{\Sigma}(t) = \Sigma_1^{-\frac{1}{2}}\big[(1-t)\Sigma_1+t(\Sigma_1^{\frac{1}{2}}\Sigma_2\Sigma_1^{\frac{1}{2}})^{\frac{1}{2}}\big]^2\Sigma_1^{-\frac{1}{2}}.
\label{eqn:interpWassersteinVar}
\end{equation}
 
There is no analytic expression for the Wasserstein metric between two GMMs. In response, a metric in the space of all GMMs, $\mathcal{GM}$, defined as
\begin{equation}
\small
D(\varrho_1, \varrho_2) = \left\{ \min_{\pi\in\Pi(\boldsymbol{\omega}_1, \boldsymbol{\omega}_2)}\sum_{i=1}^{N_1}\sum_{j=1}^{N_2}\left[W_2(g_1^i,g_2^j)\right]^2\pi(i,j)\right\}^{\frac{1}{2}},
\label{eqn:GMMWasserstein}
\end{equation}
has been proposed in \cite{chen2018optimal} as an efficient approximation to the Wasserstein metric for two arbitrary GMMs $\varrho_1 = \sum_{i=1}^{N_1}\omega_1^ig_1^i, \varrho_2 = \sum_{j=1}^{N_2}\omega_2^jg_2^j \in \mathcal{GM}$, where $\boldsymbol{\omega}_1 = [\omega_1^1,\cdots,\omega_1^{N_1}]$ and $\boldsymbol{\omega}_2 = [\omega_2^1,\cdots,\omega_2^{N_2}]$ are associated weight vectors satisfying $\sum_{i=1}^{N_1}\omega_1^i=1, \sum_{j=1}^{N_2}\omega_2^j=1$, and $\Pi(\boldsymbol{\omega}_1, \boldsymbol{\omega}_2)$ denotes the space of joint probability distributions between $\boldsymbol{\omega}_1$ and $\boldsymbol{\omega}_2$.
The geodesic path connecting $\varrho_1$ and $\varrho_2$ is given by
\begin{equation}
\small
    \hat{\varrho}_{1,2}(t) = \sum_{i,j}\pi^{\ast}(i,j)\hat{g}_{1,2}^{i,j}(t), t\in[0,1],\label{eqn:geodesic}
\end{equation}
where $\pi^{\ast}(i,j)$ denotes the optimal joint distribution, and $\hat{g}_{1,2}^{i,j}(t)$ represents the geodesic path between $g_1^i$ and $g_2^j$, which can be calculated based on \cref{eqn:interpWassersteinMean,eqn:interpWassersteinVar}.

\subsection{Conditional Value-at-Risk}
Given a PDF and a risk tolerance level $\alpha$, the CVaR computes the conditional expectation of the loss within the $\alpha$ worst-case quantile \cite{hakobyan2022distributionally}. 
The CVaR of a Gaussian random variable $\upsilon \sim \mathcal{N}(\mu,\sigma^2)$ is
\begin{equation}
\small
\text{CVaR}_{\alpha}(\upsilon) 
= \mu + \frac{\phi(\Phi^{-1}(1-\alpha))}{\alpha}\sigma,
\label{eqn:CVaR}
\end{equation}
where $\phi$ and $\Phi$ are the PDF and cumulative density function (CDF) of the standard normal distribution, respectively.

\section{Constructing Risk-Aware Gaussian Roadmap}\label{sec:WM-PRM}
The swarm's macroscopic state is represented by a time-varying PDF $\chi(t), t\in[T_0,T_f]$.
Because of the universal approximation property of GMMs, without loss of generality, we approximate $\chi(t)$ as a GMM at each time instance. The swarm's initial and target macroscopic states are modeled as two GMMs $\varrho_{T_0}$, $\varrho_{T_f}$, and the optimal PDF transport trajectory is devised in GMM space $\mathcal{GM}$, i.e., $\forall t\in[T_0,T_f], \chi(t)\in\mathcal{GM}, \chi(T_0)=\varrho_{T_0}, \chi(T_f)=\varrho_{T_f}$. 
The distance metric \cref{eqn:GMMWasserstein} and the corresponding geodesic path \cref{eqn:geodesic} in GMM space suggest that the optimal transport between two GMMs can be achieved through transport weight allocation between their respective Gaussian components. Furthermore, the geodesic path \cref{eqn:interpWassersteinMean} and \cref{eqn:interpWassersteinVar} implies that each intermediate state along the optimal transport trajectory between two Gaussian distributions retains a Gaussian distribution. Therefore, the optimal GMM trajectory $\chi(t)$ can be achieved through a set of Gaussian distribution trajectories originating from the Gaussian components of $\varrho_{T_0}$ and advancing towards those of $\varrho_{T_f}$.

To compute such trajectories, we propose to construct a risk-aware Gaussian roadmap at the macroscopic planning stage.
In contrast to conventional roadmap construction methods \cite{karaman2011sampling}, our approach samples each node as a Gaussian distribution, and determines the distance and geodesic path between each pair of nodes based on the Wasserstein metric. Furthermore, the risk measure CVaR is utilized to perform collision checking for each Gaussian node. 
This Gaussian roadmap is then used to generate GMM trajectory $\chi(t)$ between $\varrho_{T_0}$ and $\varrho_{T_f}$, which will be detailed in \cref{sec:HFSwarm}.

\subsection{Roadmap Construction}
Denote $\mathcal{G}$ as the space of two-dimensional Gaussian distributions,
where each element is represented as $g=\mathcal{N}(\boldsymbol{\mu},\Sigma)$, with the mean $\boldsymbol{\mu}\in\mathcal{W}$, and covariance matrix
\begin{equation}
\small
    \Sigma = 
\left[
  \begin{matrix}
   \sigma_1^2 & \rho\sigma_1\sigma_2 \\
   \rho\sigma_1\sigma_2 & \sigma_2^2 
  \end{matrix}
  \label{eqn:Covariance}
\right].
\end{equation}
Here $\sigma_1^2,\sigma_2^2, \rho$ denote variances and the correlation coefficient, respectively.
The distance in $\mathcal{G}$ is defined based on the Wasserstein metric $W_2:\mathcal{G}\times\mathcal{G}\rightarrow\mathbb{R}$, and the cost function is the path length in this Gaussian space.
The roadmap construction approach (\cref{alg:PRM}) is outlined in the \textsc{RoadmapConstruction} function, which takes the number of samples $n$, the connection radius $r$, the set of obstacles $\boldsymbol{\mathcal{O}}$, and a set $\mathcal{D}$ representing the set of initial and target Gaussian distributions as inputs. Initially, a node set $V$ is constructed through the combination of the set $\mathcal{D}$ as well as $n\in\mathbb{N}$ nodes generated by the function $\textsc{SampleFree}$ (Line 1). Subsequently, for each node $g\in V$, the function $\textsc{Neighbour}$ calculates the set $V_{near}$ containing all nodes located in the neighbourhood of node $g$ (Line 3). For each node $g'\in V_{near}$, the edges $(g,g')$ and $(g',g)$ are added to the edge set $E$, if the geodesic path $\hat{g}(t)$ defined based on the Wasserstein metric is a subset of the obstacle-free region in $\mathcal{G}$, which is checked by the function $\textsc{CollisionFree}$ (Line 4-7). Finally, a graph $G = (V, E)$ is constructed (Line 11).
The functions $\textsc{SampleFree}$ and $\textsc{Neighbour}$ are detailed as follows, and the functions $\textsc{inFree}$ and $\textsc{CollisionFree}$ will be detailed in \cref{sec:CVaRCollisionDetection}.

\begin{algorithm}[!t]  
        \caption{Gaussian Roadmap Construction}  
        
        \begin{algorithmic}[1] 
        \fontsize{8.5pt}{8.5pt}\selectfont
        \renewcommand{\algorithmicrequire}{ \textbf{Procedure}}
        \REQUIRE \textsc{RoadmapConstruction}$(n,r,\boldsymbol{\mathcal{O}},\mathcal{D})$
                \STATE 
                $V \gets \mathcal{D} \cup \textsc{SampleFree}(n,\boldsymbol{\mathcal{O}}); E \gets \emptyset$
                \FOR {$g \in V$}
                \STATE 
                $V_{near} \gets \textsc{Neighbour}(V\textbackslash\{g\}, g,r)$
                \FOR {$g' \in V_{near}$}
                \STATE Generate geodesic path $\hat{g}(t)$ from $g$ and $g'$ using \eqref{eqn:interpWassersteinMean} and \eqref{eqn:interpWassersteinVar}
                \IF {$\textsc{CollisionFree}(\hat{g}(t),\boldsymbol{\mathcal{O}})$}
                \STATE 
                $E \gets E \cup \{(g, g')\} \cup \{(g', g)\}$
                \ENDIF
                \ENDFOR
                \ENDFOR
                \RETURN $G = (V, E)$
        \vspace{0.2cm}
        \renewcommand{\algorithmicrequire}{ \textbf{Procedure}}
        \REQUIRE \textsc{SampleFree}$(n,\boldsymbol{\mathcal{O}})$
        \STATE $V \gets \emptyset$
        \WHILE{$| V |< n$}
        \STATE Generate Gaussian distribution $g\in\mathcal{G}$ through sampling parameter vector $\boldsymbol{v}$
        \IF{$\textsc{inFree}(g,\boldsymbol{\mathcal{O}})$}
        \STATE $V\leftarrow V\cup\{g\}$
        \ENDIF
        \ENDWHILE
        \RETURN $V$
        \vspace{0.2cm}
        \renewcommand{\algorithmicrequire}{ \textbf{Procedure}}
        \REQUIRE
        \textsc{Neighbour}$(V', g, r)$
        \STATE $V_{near}\leftarrow\emptyset$
        \FOR{$g'\in V'$}
        \IF{$W_2(g,g')\leq r$}
        \STATE $V_{near}\leftarrow V_{near}\cup\{g'\}$
        \ENDIF
        \ENDFOR
        \RETURN$V_{near}$

        \vspace{0.2cm}
        \renewcommand{\algorithmicrequire}{ \textbf{Procedure}}
        \REQUIRE \textsc{inFree}$(g,\boldsymbol{\mathcal{O}})$
                \FORALL{$\mathcal{O}\in\boldsymbol{\mathcal{O}}$}
                \STATE Compute the negation of SDF using \eqref{eqn:SDFdistribution}
                \IF{\eqref{eqn:infree} not holds}
                \RETURN\FALSE
                \ENDIF
                \ENDFOR
                \RETURN\TRUE
        \end{algorithmic}
        \label{alg:PRM}
        
\end{algorithm}

\subsubsection{\textsc{SampleFree}}
The function $\textsc{SampleFree}$ returns a set of $n \in \mathbb{N}$ nodes in free space.
We utilize five-dimensional parameter vectors $\boldsymbol{v} = [x,y,\sigma_1,\sigma_2,\rho]\in\mathbb{R}^5$ to represent Gaussian distributions in $\mathcal{G}$, with mean $[x,y]$ and covariance matrix in the form of \eqref{eqn:Covariance}. 
Whenever the number of nodes in the set $V$ is less than $n$, denoted as $|V|< n$, a node $g$ is generated through sampling a parameter vector $\boldsymbol{v}$ (Line 13-14). Various sampling strategies, including random sampling and deterministic sampling methods \cite{lavalle2006planning}, can be utilized to sample $\boldsymbol{v}$. If the Gaussian distribution $g$ is in free space, as verified by the function $\textsc{InFree}$, then it is added to the set $V$ (Line 15-17). 

\subsubsection{\textsc{Neighbour}}
The function $\textsc{Neighbour}$ takes the node $g$, connection radius $r$, and the node set $V'$ as inputs, and returns the set $V_{near}\subset V'$ containing all nodes located in the neighbourhood of
node $g$. 
For each node $g'$ in the set $V'$, if $W_2(g,g')$ is no more than the connection radius $r$,  
then $g'$ is added to the set $V_{near}$ (Line 22-24).

\begin{figure*}[!t]
\centering
\includegraphics[width=0.8\linewidth]{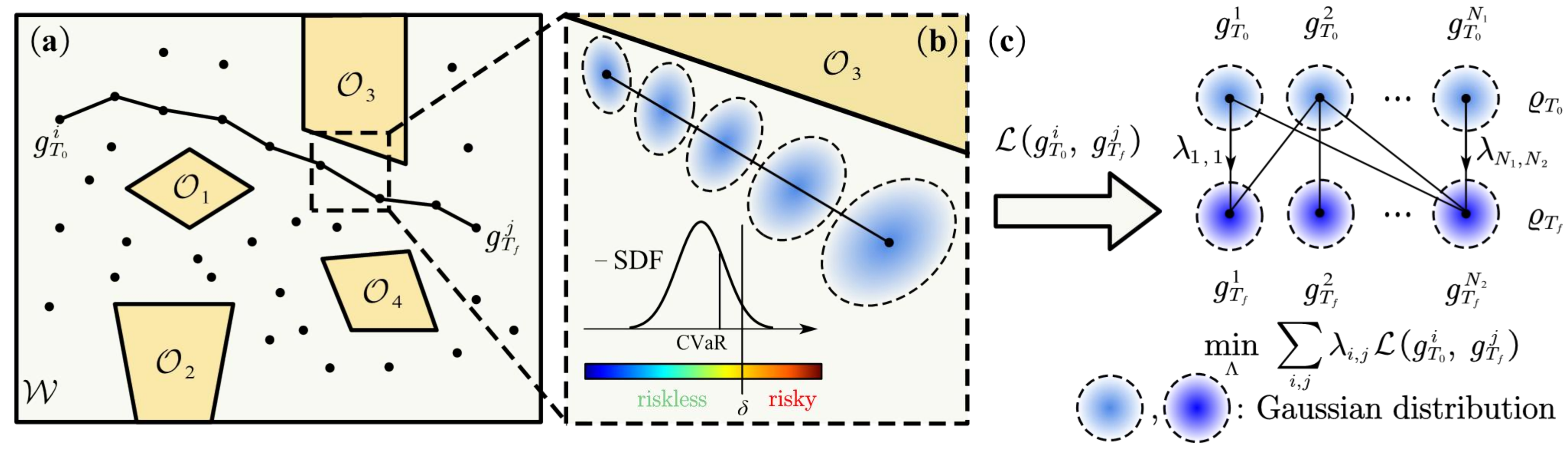}
\caption{\textbf{SwarmPRM motion planning process at the macroscopic stage.} (a) Projection of the risk-aware Gaussian roadmap onto the workspace $\mathcal{W}$, with obstacles $\mathcal{O}_1, \mathcal{O}_2, \mathcal{O}_3, \mathcal{O}_4$ colored in light orange. Each black dot represents a node in the Gaussian roadmap, corresponding to a two-dimensional Gaussian distribution. The black polyline depicts the shortest path between Gaussian distributions $g_{T_0}^i$ and $g_{T_f}^j$ on the roadmap. (b) Each polyline segment represents an edge in the Gaussian roadmap, illustrating the shortest geodesic path between two nodes based on the Wasserstein metric. The CVaR associated with each Gaussian node is required to be below the safe region threshold $\delta$. (c) Through graph search, the lowest transport cost $\mathcal{L}(g_{T_0}^i,g_{T_f}^j)$ between each pair of Gaussian distributions $(g_{T_0}^i,g_{T_f}^j)$ can be computed. The optimal GMM trajectory on the Gaussian roadmap is then calculated by solving a linear programming problem.}
\label{fig:swarmprm}
\end{figure*}

\subsection{Collision Checking Based on CVaR}\label{sec:CVaRCollisionDetection}
We utilize the risk measure CVaR to perform collision checking between each Gaussian node and the obstacles $\boldsymbol{\mathcal{O}}$ during the roadmap construction procedure.
We first derive via the linearization technique that PDF of the SDF between a Gaussian node and an obstacle can be approximated as a Gaussian distribution.
Subsequently, we utilize CVaR to constrain the conditional expectation of the SDF within the $\alpha$ worst-case quantile to reside within the safe region, where $\alpha$ denotes the risk tolerance level.

\subsubsection{SDF Linearization}
The SDF quantifies the distance between two sets $\mathcal{A}$ and $\mathcal{B}$, defined as
\begin{equation}
\small
s(\mathcal{A}, \mathcal{B}) = \left\{
\begin{aligned}
& \inf\{\Vert \boldsymbol{t} \Vert_2\ |\ \boldsymbol{t} + \mathcal{A}\cap\mathcal{B}\neq\emptyset\}, \text{if} \mathcal{A}\cap\mathcal{B} = \emptyset, \\
-& \inf\{\Vert \boldsymbol{t} \Vert_2\ |\ \boldsymbol{t} + \mathcal{A}\cap\mathcal{B} = \emptyset\}, \text{if} \mathcal{A}\cap\mathcal{B}\neq\emptyset. \\
\end{aligned}
\right.
\end{equation}
For a deterministic point $\boldsymbol{p}$ and a convex obstacle $\mathcal{O}$, we can obtain the signed distance $\hat{d}\in\mathbb{R}$, the closest points from the obstacle $\boldsymbol{p}_{\mathcal{O}}\in\mathcal{O}$, and the contact normal $\boldsymbol{n}(\boldsymbol{p},\mathcal{O}) = sgn(\hat{d})(\boldsymbol{p}_{\mathcal{O}}-\boldsymbol{p})/\Vert\boldsymbol{p}_{\mathcal{O}}-\boldsymbol{p}\Vert$ utilizing Gilbert–Johnson–Keerthi (GJK) algorithm for non-intersecting sets \cite{gilbert1988fast} and Expanding Polytope Algorithm (EPA) for overlapping shapes \cite{van2001proximity}. 

For a stochastic point $\boldsymbol{p}\sim\mathcal{N}(\boldsymbol{\mu},\Sigma)$ and a deterministic obstacle $\mathcal{O}$, the SDF can be linearized as
\begin{equation}
\small
s(\boldsymbol{p},\mathcal{O})\approx s(\boldsymbol{\mu},\mathcal{O})+\nabla s(\boldsymbol{p},\mathcal{O})\ |_{\boldsymbol{p}=\boldsymbol{\mu}}\cdot(\boldsymbol{p}-\boldsymbol{\mu}),
\label{eqn:linearize}
\end{equation}
which approximates the obstacle as a half plane with outer normal vector $-\boldsymbol{n}(\boldsymbol{\mu},\mathcal{O})$
so that $\nabla s(\boldsymbol{p},\mathcal{O})\ |_{\boldsymbol{p}=\boldsymbol{\mu}} = \boldsymbol{n}(\boldsymbol{\mu},\mathcal{O})$. It is obvious that the random variable $s(\boldsymbol{p},\mathcal{O})$ satisfies a Gaussian distribution with mean $s(\boldsymbol{\mu},\mathcal{O})$ and variance $\boldsymbol{n}^T\Sigma\boldsymbol{n}$ after linearization \cite{gao2023probabilistic,yang2023risk}, where $\boldsymbol{n}\triangleq\boldsymbol{n}(\boldsymbol{\mu},\mathcal{O})$.

\subsubsection{CVaR Collision Checking for Swarm Robots}
Given a team of robots whose macroscopic state is represented by a random variable satisfying a Gaussian distribution $\boldsymbol{p}\sim\mathcal{N}(\boldsymbol{\mu},\Sigma)$ and an obstacle $\mathcal{O}$, the probability distribution of the SDF $s(\boldsymbol{p},\mathcal{O})$ can be approximated as a Gaussian distribution. 
Define $\eta$ as the negation of the SDF, 
\begin{equation}
\small
\eta\triangleq-s(\boldsymbol{p}, \mathcal{O})\approx\eta'\sim\mathcal{N}(-s(\boldsymbol{\mu},\mathcal{O}), \boldsymbol{n}^T\Sigma\boldsymbol{n}),
\label{eqn:SDFdistribution}
\end{equation}
where a greater value of $\eta$ indicates a smaller SDF value between the robot team and the obstacle $\mathcal{O}$, and $\eta$ is approximated by the random variable $\eta'$ through \eqref{eqn:linearize}. We subsequently compute the CVaR of $\eta'$, and the collision avoidance requirement can be written as
\begin{equation}
\small
\text{CVaR}_{\alpha}(\eta, \mathcal{O})\approx \text{CVaR}_{\alpha}(\eta', \mathcal{O})\leq \delta,
\label{eqn:infree}
\end{equation}
where $\alpha$ denotes the risk tolerance level, and $\delta\leq 0$ represents the safe region threshold, constraining the conditional expectation of opposite SDF within the $\alpha$ worst-case quantile to be no more than $\delta$. 

During the construction of the Gaussian roadmap, the function $\textsc{inFree}$ serves to check whether the nodes are collision-free with respect to obstacles, i.e., whether \eqref{eqn:infree} holds for each obstacle $\mathcal{O}\in\boldsymbol{\mathcal{O}}$, as presented in \cref{alg:PRM}. For a Gaussian distribution $g$, all obstacles are queried to calculate the negation of the SDF which is a random variable following a Gaussian distribution (Line 27-28). 
Subsequently, the CVaR is calculated and the function $\textsc{inFree}$ returns \textbf{True} if \eqref{eqn:infree} is satisfied for all obstacles and \textbf{False} otherwise (Line 29-30, 33).
The function $\textsc{CollisionFree}$ (Line 6) checks whether the transition $\hat{g}(t)$ between $g$ and $g'$ is collision-free with respect to obstacles $\boldsymbol{\mathcal{O}}$, i.e., $\forall g\in \hat{g}(t), t\in[0,1], \textsc{InFree}(g,\boldsymbol{\mathcal{O}})$ returns \textbf{True}. In practical implementation, the collision checking for $\hat{g}(t)$ can be achieved approximately by performing $\textsc{InFree}$ assessments for a collection of Gaussian distributions interpolated between $g$ and $g'$ based on \cref{eqn:interpWassersteinMean} and \cref{eqn:interpWassersteinVar}.

\begin{algorithm}[!t]  
        \caption{SwarmPRM Algorithm}
        \begin{algorithmic}[1] 
        \fontsize{8.5pt}{8.5pt}\selectfont
        \REQUIRE $\varrho_{T_0}:$ the initial GMM, $\varrho_{T_f}:$ the target GMM, $\mathcal{Q}:$ initial positions of swarm robots, $\boldsymbol{\mathcal{O}}:$ environmental obstacles, $n:$ the number of nodes in the Gaussian roadmap, $r:$ the connection radius
        \ENSURE $\mathcal{T}:$ the trajectories of swarm robots 
        \STATE Obtain the set of Gaussian components $\Gamma_{T_0}, \Gamma_{T_f}$  of initial and target GMMs $\varrho_{T_0}$ and $\varrho_{T_f}$
        \STATE $G = \textsc{RoadmapConstruction}(n,r,\boldsymbol{\mathcal{O}},\Gamma_{T_0}\cup\Gamma_{T_f})$
        \FORALL{$g^i_{T_0}\in \Gamma_{T_0}, g^j_{T_f}\in \Gamma_{T_f}$}
        \STATE
        Compute the shortest path $\hat{\xi}_{i,j}(t)$ from $g^i_{T_0}$ to $g^j_{T_f}$ on $G$ using graph search
        \STATE
        Compute the transport cost $\mathcal{L}(g^i_{T_0},g^j_{T_f})$
        \ENDFOR


        \STATE Obtain $\Lambda$ by solving the LP in \eqref{P3}

        \STATE Compute $\mathcal{T}$ from $\mathcal{Q}$, $\{\hat{\xi}_{i,j}(t)\}$, and $\Lambda$ by solving \eqref{eqn:microcontrol}
        \end{algorithmic}
        \label{alg:SwarmPRM} 
\end{algorithm}

\section{SwarmPRM Approach for Hierarchical Motion Planning}\label{sec:HFSwarm}
The SwarmPRM approach (\cref{alg:SwarmPRM}) consists of the following two stages. At the macroscopic stage (Line 1-7), SwarmPRM leverages the risk-aware Gaussian roadmap constructed in \cref{sec:WM-PRM} to address the optimization problem \eqref{P} within the GMM space and generate GMM trajectory $\chi(t)$. Subsequently, the microscopic control for each robot is computed (Line 8). 
The process of SwarmPRM at the macroscopic planning stage is shown in \cref{fig:swarmprm}.

\subsection{Macroscopic Motion Planning in GMM Space}
Consider the macroscopic planning problem in GMM space $\mathcal{GM}\triangleq\{\varrho\ |\ \varrho = \sum_{i=1}^{k}\omega_ig_i,\forall g_i\in \mathcal{G}, k\in\mathbb{N}, \sum_{i=1}^{k}\omega_i=1\}$. Denote the initial and target swarm state as $\varrho_{T_0} = \sum_{i=1}^{N_1}\omega_{T_0}^ig_{T_0}^i, \varrho_{T_f} = \sum_{j=1}^{N_2}\omega_{T_f}^jg_{T_f}^j$, respectively, where $\varrho_{T_0}$ comprises $N_1$ Gaussian components $\Gamma_{T_0} = \{g_{T_0}^1,\cdots,g_{T_0}^{N_1}\}$ with weights $\boldsymbol{\omega}_{T_0} = [\omega_{T_0}^1,\cdots,\omega_{T_0}^{N_1}]$, and $\varrho_{T_f}$ consists of $N_2$ Gaussian components $\Gamma_{T_f} = \{g_{T_f}^1,\cdots,g_{T_f}^{N_2}\}$ with weights $\boldsymbol{\omega}_{T_f} = [\omega_{T_f}^1,\cdots,\omega_{T_f}^{N_2}]$. Building upon the discussions in \cref{sec:WM-PRM}, we assume the optimal transport between $\varrho_{T_0}$ and $\varrho_{T_f}$ can be achieved through a set of Gaussian trajectories $\Xi^{\ast} = \{\xi_{i,j}^{\ast}(t),t\in[T_0,T_f], i\in\{1, \cdots, N_1\}, j\in\{1, \cdots, N_2\}\}$, each $\xi_{i,j}^{\ast}(t)$ originating from the $i$th Gaussian component of $\varrho_{T_0}$ and reaching the $j$th Gaussian component of $\varrho_{T_f}$. To satisfy the normalization property of the GMM, each Gaussian trajectory $\xi_{i,j}^{\ast}(t)$ is assigned a weight $\lambda_{i,j}^{\ast}$ satisfying $\sum_{i,j}\lambda_{i,j}^{\ast}=1$, so that the optimal GMM trajectory can be derived as $\chi(t) = \sum_{i,j}\lambda_{i,j}^{\ast}\xi_{i,j}^{\ast}(t)$, and the optimal transport cost can be computed as the sum of cost along each Gaussian trajectory $\xi_{i,j}^{\ast}(t)$, which is determined by the Wasserstein metric and weighted by $\lambda_{i,j}^{\ast}$.

The SwarmPRM approach is presented in \cref{alg:SwarmPRM}. The function $\textsc{RoadmapConstruction}$ constructs a Gaussian roadmap $G = (V,E)$ including $n$ nodes $g_1,\cdots,g_n$ and the set of Gaussian components of the initial and target GMMs (Line 1-2), as detailed in \cref{sec:WM-PRM}. Through employing graph search on the Gaussian roadmap, the trajectory $\hat{\xi}_{i,j}(t)$ with the lowest transport cost $\mathcal{L}(g_{T_0}^{i},g_{T_f}^{j})$ between each pair of Gaussian distributions $(g_{T_0}^{i},g_{T_f}^{j})$ can be calculated (Line 3-6), which is an approximation to the computationally intractable optimal Gaussian trajectory $\xi_{i,j}^{\ast}(t)$. Therefore, we can provide an approximate solution to the optimization problem \eqref{P} in GMM space through computing the optimal GMM trajectory on the Gaussian roadmap. Specifically, we calculate the optimal weights $\lambda_{i,j}$ allocated to individual Gaussian trajectories $\hat{\xi}_{i,j}(t)$ to obtain the GMM trajectory $\sum_{i,j}\lambda_{i,j}\hat{\xi}_{i,j}(t)$ with minimal cost, which can be modeled as a linear programming (LP) problem (Line 7)
\begin{align} \label{P3}
\small
\min_{\Lambda}\quad & \sum_{i=1}^{N_1}\sum_{j=1}^{N_2}\lambda_{i,j}\mathcal{L}(g_{T_0}^{i},g_{T_f}^{j})\\
\textrm{s.t.}
         \quad & \sum_{i=1}^{N_1}\lambda_{i,j} = \omega_{T_f}^j, \forall j\in\{1,\cdots,N_2\}, \tag{\ref{P3}{a}} \label{P3a}\\
         \quad & \sum_{j=1}^{N_2}\lambda_{i,j} = \omega_{T_0}^i, \forall i\in\{1,\cdots,N_1\}, \tag{\ref{P3}{b}} \label{P3b}
\end{align}
where $\Lambda\triangleq\{\lambda_{i,j}, i\in\{1, \cdots, N_1\}, j\in\{1, \cdots, N_2\}\}$ denotes the transport policy.

\begin{remark}\label{rmk2}
The Gaussian trajectories planned by SwarmPRM may overlap during certain time intervals, \ie, there may exist $T_1, T_2, \hat{\xi}_{i_1, j_1},\hat{\xi}_{i_2, j_2}, T_0\leq T_1<T_2\leq T_f, \forall t\in[T_1,T_2], \hat{\xi}_{i_1, j_1}(t) = \hat{\xi}_{i_2, j_2}(t)$. The number of Gaussian components of the GMM trajectory $\chi(t)$ is thus time-varying. 
\end{remark}

\subsection{Microscopic Motion Planning}
Upon determining the optimal evolution of the macroscopic state, represented by the time-varying GMM, each robot can utilize it to compute the collision-free trajectory from the initial position to the target area (Line 8). A computationally efficient artificial potential field (APF) approach is adopted to compute the microscopic control inputs, 
\begin{equation}
\small
\boldsymbol{u}_i = -\frac{\partial (w_1U_{att}+w_2U_{rep})}{\partial \boldsymbol{u}_i}, i = 1,\cdots,N,
\label{eqn:microcontrol}
\end{equation}
where $U_{att}$, $U_{rep}$ represent the attractive and repulsive potentials, respectively, and $w_1, w_2$ are pre-defined weights representing the desired tradeoff between attractive and repulsive objectives. The detailed implementation of the APF approach can be found in \cite{zhu2021adaptive}.

\section{Simulation Results}
This section evaluates the effectiveness of the proposed approach via several simulations. A cluttered environment is used to benchmark SwarmPRM against three representative approaches both qualitatively and quantitatively. 

\subsection{Simulation Setup}
\begin{figure*}[!t]
\centering
\includegraphics[width=0.95\linewidth]{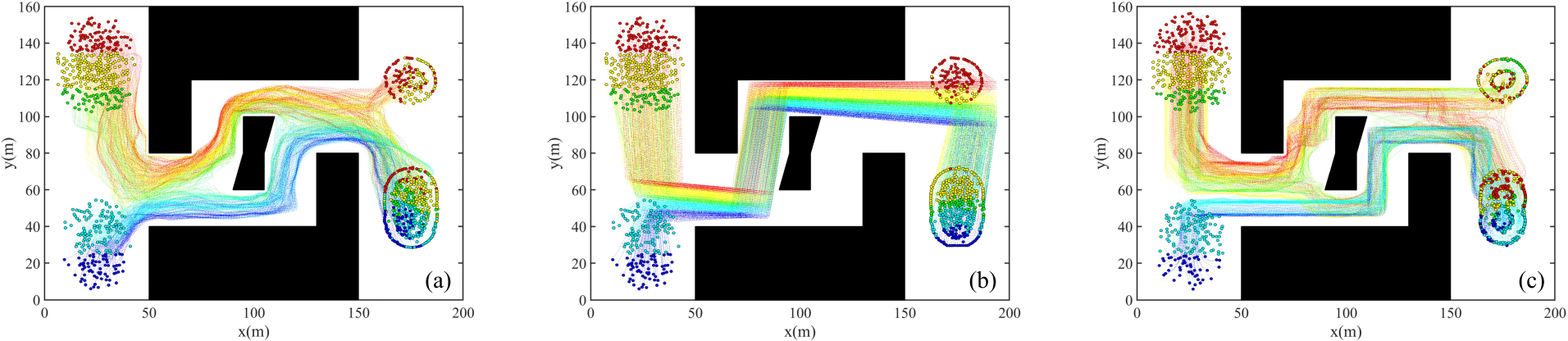}
\caption{\textbf{Qualitative simulation results.} The trajectories of the swarm robotic system comprised of $N=500$ robots obtained by
(a) SwarmPRM, (b) Formation control, and (c) ADOC. The initial and final positions are represented by colored circles on the left and right sides of each subfigure, respectively, while the obstacles are depicted in black color.}
\label{fig:Graph_allscenes}
\end{figure*}

A representative two-dimensional environments containing $3$ nonconvex obstacles (\cref{fig:Graph_allscenes}) is designed to evaluate the performance of our approach both qualitatively and
quantitatively. Swarm robotic systems consisting of $N = 20, 40, 100, 500$ robots are tasked with navigating from a given initial distribution $\varrho_{T_0}=\sum_{i=1}^{N_1}\omega_{T_0}^ig_{T_0}^i$ to a target distribution $\varrho_{T_f}=\sum_{j=1}^{N_2}\omega_{T_f}^jg_{T_f}^j$, while avoiding collisions with obstacles in $\mathcal{W} = [0, W] \times [0, H]$, where $W = 200$m, $H = 160$m, $N_1 = 4$, $N_2 = 3$, $\omega_{T_0}^1=\frac{1}{4}$, $\omega_{T_0}^2=\frac{3}{8}$, $\omega_{T_0}^3=\frac{3}{16}$, $\omega_{T_0}^4=\frac{3}{16}$, $\omega_{T_f}^1=\frac{1}{4}$, $\omega_{T_f}^2=\frac{3}{8}$, $\omega_{T_f}^3=\frac{3}{8}$. All Gaussian components of $\varrho_{T_0}$ and $\varrho_{T_f}$ share the same covariance matrix, $100I_2$, where $I_2$ denotes the second-order identity matrix. The means of $g_{T_0}^i, i\in\{1, 2, 3, 4\}$ and $g_{T_f}^j, j\in\{1, 2, 3\}$ are $[25,20]$, $[25,40]$, $[25,120]$, $[25,140]$, and $[175,40]$, $[175,60]$, $[175,120]$, respectively.
Each robot is of circular shape with a radius of $0.2$m and is characterized by  single-integrator dynamics
\begin{equation}
\small
    \dot{\boldsymbol{x}}_i(t) = \boldsymbol{u}_i(t), \boldsymbol{x}_{i}(T_0) = \boldsymbol{q}_{i},\forall i \in\{1,\cdots,N\},
\end{equation}
where $\boldsymbol{x}_i$ denotes the robot position, and the control input $\boldsymbol{u}_i$ is a vector of linear velocities in the $x$- and $y$-directions. 
When constructing the Gaussian roadmap, we set the number of samples $n = 500$, and the connection radius $r = 20$. 

We compare our approach with three state-of-the-art methods including multi-robots formation control \cite{alonso2017multi}, discrete RRT$^\ast$ \cite{shome2020drrt}, and adaptive distributed optimal control \cite{zhu2021adaptive}, denoted as Formation control, dRRT$^\ast$, and ADOC, respectively. 
The performance can be assessed by the computational time $T_{sol}$ (min) and the average trajectory length $\overline{D}$ (m), 
\begin{equation}
\small
    \overline{D} = \frac{1}{N}\sum_{i=1}^N\sum_{\tau=0}^{\kappa-1}\Vert\boldsymbol{x}_i[T_0+(\tau+1)\Delta t]-\boldsymbol{x}_i(T_0+\tau\Delta t)\Vert_2,
\end{equation}
where $\kappa = \frac{T_f-T_0}{\Delta t}$. The simulations are implemented using MATLAB code, and executed on a desktop with Intel Core i7-14700KF CPU@3.40 GHz and 32GB RAM. In practical implementation, we integrate the uniform sampler with a Gaussian sampler \cite{boor1999gaussian} to achieve a better coverage of the difficult area in the collision-free space. Furthermore, to avoid excessively high local densities in the swarm macroscopic state which adds difficulty to microscopic planning, we impose an upper limit on the probability density of the GMM $\chi(t)$ solved by the LP problem \eqref{P3}, which can be formulated as a minimum-cost flow problem \cite{ahuja1988network}.

\begin{table}[!t]
\caption{Quantitative Simulation Results}
\label{tab:env1}
\begin{center}
\resizebox{\linewidth}{!}{
\begin{tabular}{|c|c|c|c|c|}
\hline
\textbf{$N$}& \textbf{SwarmPRM}& \textbf{Formation control}& \textbf{dRRT$^\ast$}& \textbf{ADOC} \\
\hline
$20$& $3.4(0.05)\ /\ 241.6(1.31)$& $\boldsymbol{2.2}(0.30)\ /\ 305.2(14.82)$& $\boldsymbol{229.8}(18.11)$& $6.4\ /\ 295.6$\\
\hline
$40$& $\boldsymbol{3.5}(0.09)\ /\ \boldsymbol{240.7}(1.36)$& $5.3(0.53)\ /\ 312.5(9.91)$& $408.1(25.91)$& $6.9\ /\ 292.1$\\
\hline
$100$& $\boldsymbol{3.9}(0.14)\ /\ \boldsymbol{237.1}(1.93)$& $8.6(0.41)\ /\ 304.0(5.39)$& $-$& $7.5\ /\ 282.5$\\
\hline
$500$& $\boldsymbol{7.1}(0.50)\ /\ \boldsymbol{236.1}(1.98)$& $54.6(5.48)\ /\ 309.3(14.04)$& $-$& $13.2\ /\ 279.0$\\
\hline
\end{tabular}}
\label{tab1}
\end{center}
\end{table}

\subsection{Results and Analyses}
The quantitative results are shown in \cref{tab:env1}, with the values before and after the slash representing $T_{sol}$ and $\overline{D}$, respectively. We set the computational time limit of dRRT$^\ast$ to be $90$ minutes, i.e., $T_{sol}=90$, and record $\overline{D}$ when dRRT$^\ast$ terminates. It should be noticed that SwarmPRM, Formation control and dRRT$^\ast$ involve sampling procedures, and the corresponding simulation results are mean values obtained from five random simulations, with the standard deviations indicated in parentheses.
In comparison with ADOC, which plans the time-varying GMM using a predetermined set of collocation Gaussian components, our approach demonstrates superior performance in both metrics, showcasing the efficacy of the sampling-based method on generating GMM trajectories with lower costs within shorter computational time. 
Regarding Formation control, the global planner necessitates adhering to the formation for the entirety of the swarm, leading to an increased average trajectory length compared to other methods. Moreover, the local motion planning entails solving distributed nonlinear optimizations for each robot at high frequency to avoid collisions, resulting in a rapid increase in the overall computational time $T_{sol}$ as the swarm size grows, which exceeds that of SwarmPRM in scenarios where $N = 40, 100, 500$. In comparison with dRRT$^\ast$, our approach demonstrates great computational efficiency and computes solutions with lower $\overline{D}$ when the swarm size is $40$, thanks to the hierarchical planning strategy. 
The dRRT$^\ast$ computes solutions with the lowest $\overline{D}$ when the swarm size is $20$. However, the solution quality of dRRT$^\ast$ decreases sharply when the swarm size is $40$ due to the substantial computational burden. For swarm robotic systems consisting of $100$ or $500$ robots, dRRT$^\ast$ cannot find any solution within the allotted computational time. 

The qualitative results in \cref{fig:Graph_allscenes} show the trajectories of the swarm robotic system comprising $500$ robots planned by SwarmPRM, Formation control, and ADOC. All approaches successfully plan a collision-free trajectory for each robot with respect to obstacles and other robots. The SwarmPRM and ADOC allow the swarm to split and merge, resulting in shorter trajectory lengths than Formation Control, which requires the maintenance of a formation for all the robots. 

\section{Conclusion}
We develop the SwarmPRM motion planning approach, offering a new perspective on developing sampling-based motion planning methods for large-scale swarm robotic systems. 
We propose to construct a risk-aware Gaussian roadmap, where each node represents a Gaussian distribution, and CVaR is incorporated for collision checking. 
We then formulate a linear programming problem to calculate the optimal GMM trajectory on the roadmap. 
Simulation results validate the computational efficiency, scalability, risk awareness, and trajectory flexibility of SwarmPRM. 
Future work includes extension to three-dimensional environments and implementation of SwarmPRM on real robotic platforms.

{
\small
\bibliography{ref}
\bibliographystyle{IEEEtran}
}

\end{document}

%% file: configs.tex
\let\labelindent\relax
\usepackage{graphicx} \graphicspath{ {figures/} }
\usepackage{amsmath,amssymb,mathabx,amsbsy,mathtools,etoolbox}
\usepackage[utf8]{inputenc} 
\usepackage[T1]{fontenc}    

\usepackage{algorithmic}
\usepackage[ruled,linesnumbered]{algorithm2e}
\usepackage{acronym}
\usepackage{enumitem}
\usepackage[breaklinks,colorlinks,linkcolor=black]{hyperref}
\usepackage{balance}
\usepackage{xspace}
\usepackage{setspace}
\usepackage[skip=3pt,font=small]{subcaption}
\usepackage[skip=3pt,font=small]{caption}
\usepackage[dvipsnames,svgnames,x11names]{xcolor}
\usepackage[capitalise,nameinlink]{cleveref}
\usepackage{booktabs,tabularx,colortbl,multirow,multicol,array,makecell}
\usepackage{pifont}
\usepackage{cite}
\usepackage{nicematrix}

\makeatletter
\DeclareRobustCommand\onedot{\futurelet\@let@token\@onedot}
\def\@onedot{\ifx\@let@token.\else.\null\fi\xspace}

\def\ie{\emph{i.e}\onedot}

\makeatother

\makeatletter
\def\BState{\State\hskip-\ALG@thistlm}
\makeatother

\makeatletter
\renewcommand{\paragraph}{%
  \@startsection{paragraph}{4}%
  {\z@}{0ex \@plus 0ex \@minus 0ex}{-1em}%
  {\hskip\parindent\normalfont\normalsize\bfseries}%
}
\makeatother

\crefname{algorithm}{Alg.}{Algs.}
\Crefname{algocf}{Algorithm}{Algorithms}
\crefname{section}{Sec.}{Secs.}
\Crefname{section}{Section}{Sections}
\crefname{table}{Tab.}{Tabs.}
\Crefname{table}{Table}{Tables}
\crefname{figure}{Fig.}{Fig.}
\Crefname{figure}{Figure}{Figure}

\definecolor{gblue}{HTML}{4285F4}
\definecolor{gred}{HTML}{DB4437}
\definecolor{ggreen}{HTML}{0F9D58}

\definecolor{mygray}{gray}{.92}

\acrodef{qp}[QP]{Quadratic Programming}
\acrodef{fd}[FD]{Force Decomposition}
\acrodef{ros}[ROS]{Robot Operating System}
\acrodef{uav}[UAV]{Unmanned Aerial Vehicle}
\acrodef{dof}[DoF]{Degree-of-freedom}
\acrodef{com}[CoM]{Center-of-Mass}
\acrodef{ilc}[ILC]{Iterative Learning Control}
\acrodef{siso}[SISO]{single-input-single-output} 
\acrodef{zpetc}[ZPETC]{zero-phase-error tracking controller}
\acrodef{mimo}[MIMO]{multi-input-multi-output}

\let\oldnl\nl
\newcommand{\nonl}{\renewcommand{\nl}{\let\nl\oldnl}}%
\makeatother